# AUTOMATION: AN ESSENTIAL COMPONENT OF ETHICAL AI?

Vivek Nallur
*School of Computer Science, University College Dublin, Ireland*
*vivek.nallur@ucd.ie*

Martin Lloyd, Siani Pearson
*Just Algorithm Action Group (JAAG), UK*
*siani.pearson@jaag.org.uk*

**ABSTRACT**

Ethics is sometimes considered to be too abstract to be meaningfully implemented in artificial intelligence (AI). In this paper, we reflect on other aspects of computing that were previously considered to be very abstract. Yet, these are now accepted as being done very well by computers. These tasks have ranged from multiple aspects of software engineering to mathematics to conversation in natural language with humans. This was done by automating the simplest possible step and then building on it to perform more complex tasks. We wonder if ethical AI might be similarly achieved and advocate the process of automation as key step in making AI take ethical decisions. The key contribution of this paper is to reflect on how automation was introduced into domains previously considered too abstract for computers.

## 1. INTRODUCTION

In this paper we examine ways in which abstract concepts, previously thought to be the preserve of human thought have been increasingly subject to automation. We take a journey through computing capabilities step by step towards complexity. We reflect on these examples, painting a vision in Section 5 of how automation in the ethical field might progress in future and offer a suggestion for what the first step could be.

## 2. AUTOMATING THE SMALLEST VIABLE SE PRACTICE

Software engineering (SE) has advocated the use of testing, and formalised code reviews as a means of increasing software quality for a long time (Fagan, 1976). However, as engineers who created and deployed software in 1980-90s can attest, even though it was a good idea, there was very little implementation. This is because writing good, elegant code was (and still is) considered to be an art form. This implies two things: a) some people (the artists) are a cut above the others; b) there are no rules that can make all code elegant (it's not engineering, it's an art). SE practices therefore focussed mainly on understanding the customers' requirements and being able to produce designs and code that met those requirements. This changed drastically with the introduction of automation.

The introduction of automated unit-testing, via SUnit, and subsequently jUnit by Kent Beck (Beck, 1997) allowed developers to add tests to their code, as they were writing the functionality, and not as an after-thought. This simple idea of automatically testing a small piece of functionality with well-defined entry and exit criteria increased the amount of testing being done by developers drastically. Critically, now unit-testing is seen as a first-class programming artifact that is checked into version-control, along with the code that it tests. Development methodologies such as TestDrivenDevelopment and ExtremeProgramming use unit tests as a critical part of the software development process. Automation of unit-testing has been almost universally accepted as a good practice, so much so that several programming languages, such as Go, Matlab, Python, Ruby, Rust, etc. support unit-testing directly within the programming language, while almost all widely used programming languages have extremely good testing libraries/frameworks. In systems that need testing for certification as per the IEC 61508 systems safety standard, unit-testing of all functions and all possible branches is mandated. This is only possible in modern systems through the introduction of automated testing.

Code-reviews are another example of a software engineering practice that was considered as requiring abstract skills. The practice of other team-members critiquing changes to the codebase is acknowledged to be a good way to catch bugs, provide feedback, and increase software quality. However, this requires software teams to invest time and energy in understanding the *intent* of the code, as well as its impact on the rest of the system, even parts that have not been built yet. Again, we see the recurrent pattern of the ideal activity taking a lot of time and human expertise to do well. This leads to a high cost of performing the activity, and a consequent reluctance within teams and organisations to invest in it, which leads to poor software quality (McIntosh et al, 2016). While code-review is a complex, cognitive task, some of the burden was eased with the introduction of automation. The use of automated style-checkers, linters, static analysis tools in software systems reduces the burden of the human reviewer who can now focus only on the most difficult parts of bug-finding, sharing design intent and knowledge transfer (Bacchelli and Bird, 2013).

Modern software systems are often composed of heterogenous micro-services, built by multiple teams from different enterprises. Composing these into a cohesive application that can service a user's needs requires human thought and ingenuity. Even in this domain, there are attempts to automatically create applications built from micro-services (Cabrera, C et al. 2017) and repair/adapt them dynamically in the face of perturbances (Song, H. et al. 2015; Nallur, V.et al. 2016). The field of software engineering has embraced automation as a way of making complex, cognitive tasks readily repeatable, as well as reliable.

## 3. AUTOMATING MATHEMATICS

Up to about thirty years ago many mathematicians would claim that there is something special about human insight, and that computers could never come up with the kind of 'eureka' moments that can occur in finding a mathematical proof. A formal mathematical proof is a finite sequence of sentences, each of which is an axiom, an assumption that follows from the earlier sentences by a rule of inference. Verifying a formal proof is purely mechanical, but it is often hard to formulate rules to find the proofs.

There are different ways in which AI has been used to contribute to the automation of mathematics. Some of these produce an interactive system in order to suggest ideas to the user, to allow description and manipulation of proofs and also to automate trivial steps which would be tedious to do by hand: for example, verification theorem-provers like Isabelle (Paulson, 1994). Others take the ambitious goal of creating an 'artificial mathematician'. This can involve detection of general patterns of proof and provision of maximum possible automation of proof. A computer following a purely algorithmic approach can consider many more approaches than a human, and look for consequences further ahead before deciding how promising each potential approach would be. Once faster computers became available, initial inefficient symbol-crunching approaches were improved by normalisation procedures (Gilmore, 1970), and by further refinement of standard procedures in the form of resolution (Robinson, 1965), then by advances in high-level tactical reasoning such as CLAM (Bundy et al, 1990).

Some very large mathematical proofs have been verified using interactive theorem provers, where humans and machines collaborate. Long computer-assisted proofs have been produced, including for the four-colour problem, Kepler conjecture and Boolean Pythagorean triples problem. The automated theorem prover EPQ in 1996 proved the Robbins conjecture. In summary, there is much progress in tackling what used to be thought as the human preserve of mathematical intuition.

## 4. AUTOMATING CONVERSATION

The Turing Test (originally called the Imitation Game) was proposed as a test of a machine's intelligence (Turing, 1950): a human evaluator observes a natural language conversation between a machine and a human via a textual/online communication channel (to omit other clues) and has to decide whether s/he can distinguish which entity is the machine. In 1966, Joseph Weizenbaum's program Eliza was the first to pick up the challenge, being a simple keyword-based program that mimicked a human psychiatrist (Weizenbaum, 1966). This was the start of natural language processing (NLP), which progressed from symbolic to statistical and neural approaches (Zakaria Kurdi, 2017). In 1971, *Parry* a computer program that simulated a person with schizophrenia, was the first program considered to have passed a modified Turing Test, (Colby, 1972). Various other developments followed until 2010, when Apple launched Siri, after which virtual assistants and chatbots have widely expanded in development and application.

Many of these chatbots use AI, thereby being able to respond to the context and structure of the information being provided and provide a more realistic conversation experience. They may use: Machine Learning (*ML)* to identify patterns in user input, make decisions and learn from past conversation; *NLP* to understand the meaning and context of natural language, deal with human error in the input and respond naturally; *sentiment analysis* to help a chatbot understand users' emotions. The current state-of-the-art is is OpenAI's GPT3 (Generative Pretrained Transformer), which uses ML to automate a variety of NLP tasks like text translation and answering questions (Floridi and Chiriatti, 2020).

## 5. AUTOMATION IN ETHICS

Ethical behaviour from AI-assisted machines has been in the news, mostly due to problematic cases that have come to light, such as bias in judicial decision-making[1], bias in recruitment[2], chatbots that started to use racist language[1], etc. These incidents have highlighted a weakness of the current techniques employed in AI-based systems, i.e., the tendency of the machines to reproduce historical bias or other infirmities in the learning data. More fundamental questions have been asked about whether any AI could ever be trusted with decisions that impact human life (Krishnan, 2009; Misselhorn, 2018). We think it is important to distinguish between ethics *inside* AI-based systems, ethics of businesses *developing* AI-based systems, and ethics of AI *deployments*. Although it is important for all three aspects to be ethical, we posit that the ethical problems involved in these three aspects may be solved using different techniques. For instance, the question of ethics *inside* AI seems more abstract, and current efforts fall short of expectations (Nallur, 2020). On the other hand, ethics of people *developing* AI and the ethics of *deploying* AI may be more amenable to being solved using currently available techniques. There are already some techniques that have been recommended to develop AI in an ethical manner (Anderson *et al*, 2019; Gebru *et al*, 2020).

Deployment of AI-based systems requires analysis of development techniques, social contexts, as well as the deploying organisation's own practices. This is currently done through an exhaustive investigation of physical and digital artefacts by human beings. In our opinion, there will be increasing efforts to automatically identify digital artefacts that could be used to infer ethically problematic practices. Initially, these might be in the form of a bot that is limited to generating reports, which are then evaluated by a human. However, in the same manner as automated testing, theorem proving, and NLP, we might be able to automate small tasks that build on each other to perform a complex operation. This may result in more sophisticated tools that may eventually be able to perform deeper analysis on a company's artefacts to infer ethical values, in the same manner that the ISO 9000 series of certifications allows us to infer product quality from process quality. Correctness-by-Construction (White, 2017) recommends this very same philosophy: By taking small steps, each step-by-step verification activity is correspondingly small and manageable. By using precise notations, each verification step can be trustworthy and amenable to automation, and there are no gaps for defects to hide in. By doing tool-supported verification, we reduce the element of human error.

**Arguments against Automating Ethical Reasoning:** Care needs to be taken that the automated process itself does not introduce new ethical problems, including lack of transparency and human control (Hagendorff, 2020; IEEE, 2019). Another concern is that the presence of automation tends to make humans shed their cognitive engagement. This has been flagged by some ethicists as potentially leading to a situation where we abrogate our responsibility to act in an ethical manner. A third concern is how intrusive such technologies would need to be, and whether such intrusion is warranted or trusted.

---

[1] https://www.propublica.org/article/how-we-analyzed-the-compas-recidivism-algorithm

[2] https://www.reuters.com/article/us-amazon-com-jobs-automation-insight-idUSKCN1MK08G

# 6. CONCLUSIONS

In this paper, we have considered several tasks that were previously considered to require human ingenuity, and capacity for abstract thought. We have also outlined how these domains have seen progress over a few decades such that current tools/systems are perceived to be functioning at human levels of competence or better. There is no known physical/informational limit that precludes the domain of ethical evaluation of actions from such efforts. We would speculate that simpler tasks such as checking whether datasets have been de-biased would be the first milestone in this endeavour, with other practices following suit. Automation thus functions as a marker of human ingenuity, where we are able to *dissociate* the task from the intelligence required to do it, by breaking it down into simpler objectively evaluated parts.